\documentclass[runningheads,a4paper]{llncs}
\usepackage{paralist} 
\usepackage{amsmath}
\usepackage{amssymb,bm,amsfonts,mathrsfs,amsbsy,amssymb}
\usepackage{graphicx}
\usepackage{multirow} 
\usepackage{enumerate} 
\usepackage{caption}
\usepackage{url}
\usepackage{verbatim}
\usepackage{multirow}

\newcommand{\norm}[1]{\left\lVert#1\right\rVert}

\begin{document}

\mainmatter  % start of an individual contribution

\title{Improved Knowledge Base Completion by Path-Augmented TransR Model}
\titlerunning{Path-Augmented TransR Model}


\author{Wenhao Huang, Ge Li, Zhi Jin}
\authorrunning{Wenhao Huang, Ge Li, Zhi Jin}

\institute{Key  Laboratory of High Confidence Software Technologies (Peking University),\\
	Ministry of Education, China\\
	Software Institute, Peking University, China\\
	\texttt{wenhao.huang@pku.edu.cn}\\
	\texttt{\{lige,zhijin\}@sei.pku.edu.cn}
}


%\toctitle{Lecture Notes in Computer Science}
%\tocauthor{Authors' Instructions}
\maketitle

\renewcommand{\baselinestretch}{1.05}

\begin{abstract}
	
Knowledge base completion aims to infer new relations from existing information. In this paper, we propose path-augmented TransR (PTransR) model to improve the accuracy of link prediction. In our approach, we base PTransR model on TransR, which is the best one-hop model at present. Then we regularize TransR with information of relation paths. In our experiment, we evaluate PTransR on the task of entity prediction. Experimental results show that PTransR outperforms previous models.

\end{abstract}

\section{Introduction}\label{sec:intro}

Large scale knowledge bases such as WordNet~\cite{wordnet} and FreeBase~\cite{freebase} are important resources for natural language processing (NLP) applications like web searching~\cite{websearch}, automatic question answering systems~\cite{QA}, and even medical informatics~\cite{gene}. Formally, a \textit{knowledge base} is a dataset containing triples of two entities and their relation. A triplet $(h,r,t)$, for example, indicates that the \textit{head entity} $h$ and the \textit{tail entity} $t$ have a relation $r$. Despite massive triplets a knowledge base contains, evidence in the literature suggests that existing knowledge bases are far from complete~\cite{openIE,TransH}. %TODO

In the past decades, researchers have proposed various methods to automatically construct or populate knowledge bases from plain texts~\cite{openIE,SDP-LSTM}, semi-structured data on the Web~\cite{webIE,wiki}, etc.
Recently, studies show that embedding the entities and relations of a knowledge base into a continuous vector space is an effective way to integrate the global information in the existing knowledge base and to predict missing triplets without using external resources (i.e., additional text or tables)~\cite{TransE,TransH,TransR,liang,PTransE}.

Bordes~et al.~\cite{TransE} propose the TransE approach, which \underline{trans}lates entities' \underline{e}mbeddings by that of a relation, to model knowledge bases. That is to say, the relation between two entities can be represented as a vector offset, similar to word analogy tasks for word embeddings~\cite{analogy} and sentence relation classification by sentence embeddings~\cite{SNLI}. For one-to-many, many-to-one, and many-to-many relations, however, such straightforward vector offset does not make much sense. Considering a head entity \texttt{China} and the \texttt{country-city} relation, we can think of multiple plausible tail entities like \texttt{Beijing}, \texttt{Tianjin}, and \texttt{Shanghai}. These entities cannot be captured in the same time by translating the head entity and relation embeddings.
Therefore, researchers propose to map entities to a new space where embedding translation is computed, resulting in TransH \cite{TransH}, TransR \cite{TransR}, and other variants. Among the above approaches, TransR achieves the highest performance on established benchmarks.

One shortcoming of the above methods is that only the direct relation (i.e., one-hop relation) between two entities is considered. In a knowledge base, some entities and relations only appear a few times; they suffer from the problem of data sparsity during training. Fortunately, the problem can be alleviated by using multi-hop information in a knowledge base. Guu et al.~\cite{liang} present a random walk approach to sample entities with composited relations. Likewise, Lin et al.~\cite{PTransE} propose a path-augmenting approach that uses multi-hop relations between two entities to regularize the direct relation between the same entity pair. Their experiments show the \underline{p}ath-augmented TransE model
(denoted as PTransE) outperforms the one-hop TransE model.

In this paper, we are curious whether we can combine the worlds, i.e., whether the path-augmenting technique is also useful to a better one-hop ``base'' model. Therefore, we propose to leverage TransR \cite{TransR} as our cornerstone, but enhance it with \underline{p}ath information as in \cite{PTransE}, resulting a new variant, PTransR. We evaluate our model on the FreeBase dataset.
Experimental results show that modeling relation paths is beneficial to the base model TransR, and that PTransR also outperforms PTransE in entity prediction. In this way, we achieve the state-of-the-art link prediction performance in the category that uses only the knowledge base itself (i.e., without additional textual information).

The rest of this paper is organized as follows. In Section~\ref{sec:model}, we describe the base model TransR and then discuss the path-augmented variant PTransR. In Section~\ref{sec:exp}, we compare our PTransR model with other baselines in an entity prediction experiment; we also have in-depth analysis regarding different groups of relations, namely $1$-to-$1$, $1$-to-$n$, $n$-to-$1$, and $n$-to-$n$ relations. In Section~\ref{sec:relatedwork}, we briefly review previous work in information extraction. Finally, we conclude our paper in Section~\ref{sec:conclusion}.

\section{Our Approach}\label{sec:model}

In this section, we present our PTransR model in detail. In Subsection~\ref{ss:TransR}, we introduce the TransE model and explain how TransR overcomes the weakness of TransE. Then, we augment TransR model with path information in Subsection~\ref{ss:PTransR}.

\subsection{Base Model: TransR}\label{ss:TransR}

As said in Section~\ref{sec:intro}, embedding entities and their relation into vector spaces can effectively exploit internal structures that a knowledge base contains, and thus is helpful in predicting missing triplets without using additional texts.

The first model in such research direction is TransE~\cite{TransE}. It embeds entities and their relation in a same low-dimensional vector space; the two entities' embeddings are translated by a relation embedding, which can be viewed as an offset vector. In other words, for a triplet $(h, r, t)$, we would like $\textbf{h}+\textbf{r} \approx \textbf{t}$. (Here, bold letters refer to the embeddings of head/tail entities and the relation.) The plausibility of a triplet ($h,r,t$) is then evaluated by a scoring function
\begin{equation}
f_r(\textbf{h},\textbf{t}) \triangleq \norm{\textbf{h}+\textbf{r}-\textbf{t}},
\end{equation}
where $\|\cdot\|$ denotes either $\ell_1$-norm or $\ell_2$-norm. $f_r(\textbf{h},\textbf{t})$ is expected to be small if ($h,r,t$) is a positve triplet. 

To further analyze the performance of TransE, Bordes et al.~\cite{TransE} divide relations into four groups, namely $1$-to-$1$, $1$-to-$n$, $n$-to-$1$, and $n$-to-$n$, according to the mapping properties of a relataion. For example, \texttt{country-city} is a $1$-to-$n$ relation, because a country may have multiple cities, but a city belongs to only one country. The weakness of TransE is that entity embeddings on the \texttt{many} side tend to be close to each other, which is the result of expecting $f_r(\textbf{h},\textbf{t})$ to be small for all positive triplets. Therefore, it is hard for TransE to distinguish among the entities which are on the \texttt{many} side.

To solve the above problem, TransR~\cite{TransR} embeds entities and relations into two separate spaces: the \textit{entity space} and the \textit{relation space}. It uses relation-specific matrices $\mathbf{M}_r$ to map an entity from its own space to the relation space, given by $ \mathbf{h}_{r} = \mathbf{M}_{r}\mathbf{h} $ and $\mathbf{t}_r=\mathbf{M}_r \mathbf{t}$, so that translation can be accomplished by regarding relation embedding as an offset vector, i.e., 
$\textbf{h}_{r}+\textbf{r}\approx\textbf{t}_{r}$. To achieve this goal, TransR defines the scoring function as 
\begin{equation}
f_r(\textbf{h},\textbf{t})\triangleq\norm{\textbf{M}_{r}\textbf{h}+\textbf{r}-\textbf{M}_{r}\textbf{t}}_2^2 .\label{eqn:TransR}
\end{equation}

To train the model, we shall generate negative samples and use the hinge loss. The overall cost function of the TransR model is
\begin{equation}
\mathcal{L}_\text{TransR} = \sum_{(h,r,t)\in S} \sum_{(h',r,t')\in S'} \max\big\{0, \gamma + f_r(\textbf{h},\textbf{t})-f_r(\textbf{h}',\textbf{t}')\big\},
\end{equation}
where negative samples are constructed as $$S' = \left\{(h',r,t)|(h',r,t)\notin S\right\} \bigcup \left\{(h,r,t')|(h,r,t')\notin S\right\}.$$ %Entity and relation embeddings are obtained by minimizing $\mathcal{L}_{TransR}$

Results in link prediction show that TransR outperforms other models on established benchmarks, indicating TransR is the best one-hop model at present. However, TransR fails to utilize the rich path information, which will be dealt with in the following subsection.

\subsection{Path-Augmented TransR: PTransR}\label{ss:PTransR}

\begin{figure}[!t]
	\centering
	\includegraphics[width=.6\textwidth]{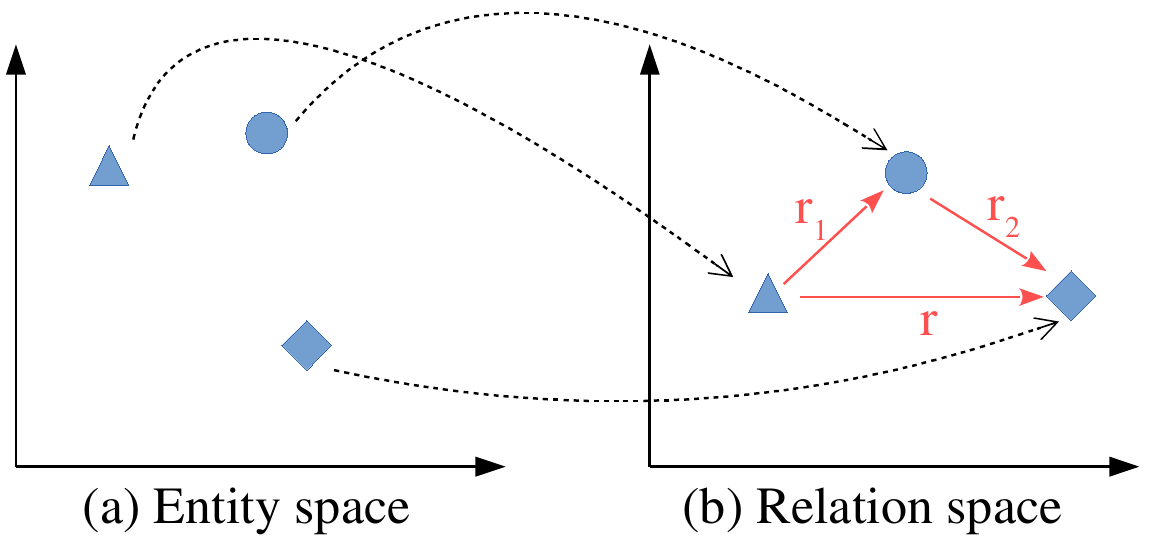}
	\caption{An illustration of the PTransR model.}
	\label{fig:PTransR}
\end{figure}

Using path information to regularize one-hop models can be beneficial~\cite{PTransE,liang}. Here, we adopt the path modeling method in PTransE, and extend the TransR model to path-augmented TransR (denoted as PTransR).

A relation path is a set of relations that connect head entity and tail entity in succession. An $n$-hop relation path from $h$ to $t$ is defined as
$p = \{r_1,r_2,\cdots,r_n\}$, satisfying  $h\xrightarrow{r_1}e_1\xrightarrow{r_2}\cdots\xrightarrow{r_n}t$. If $n = 1$, then $p=r_1$ is a direct (1-hop) relation.
To enhance TransR model with multi-hop information, we follow the treatment in PTransE~\cite{PTransE} and represent a relation path as an embedding vector by additive compositional methods. Then such multi-hop information is used to regularize one-hop direct relation between the same entity pair. A reliability score is computed to address the strength of regularization by a particular path. Details are described as follows.

To compute the representation of a relation path $p$ that composites primitive relations $r_1, r_2,\cdots, r_n$, i.e., $p = r_1 \circ r_2 \circ \cdots \circ r_n$ (where $\circ$ denotes the composition operation), we add the embeddings of these primitive relations, given by
\begin{equation}
\mathbf{p} = \mathbf{r}_{1}+\mathbf{r}_{2}+\cdots+\mathbf{r}_{n},
\end{equation}
where bold letters denote the vector of a relation or a path.

The choice of addition as the composition operation is reasonable, because the vector representation of path $p$ should be close to that of direct relation $r$ if it is likely to infer $r$ from $p$. For example, the representation of path $\xrightarrow{father} \xrightarrow{mother}$ is expected to be close to that of direct relation $\xrightarrow{grandmother}$.

\begin{figure}[!t]
	\centering
	\includegraphics[width=.6\textwidth]{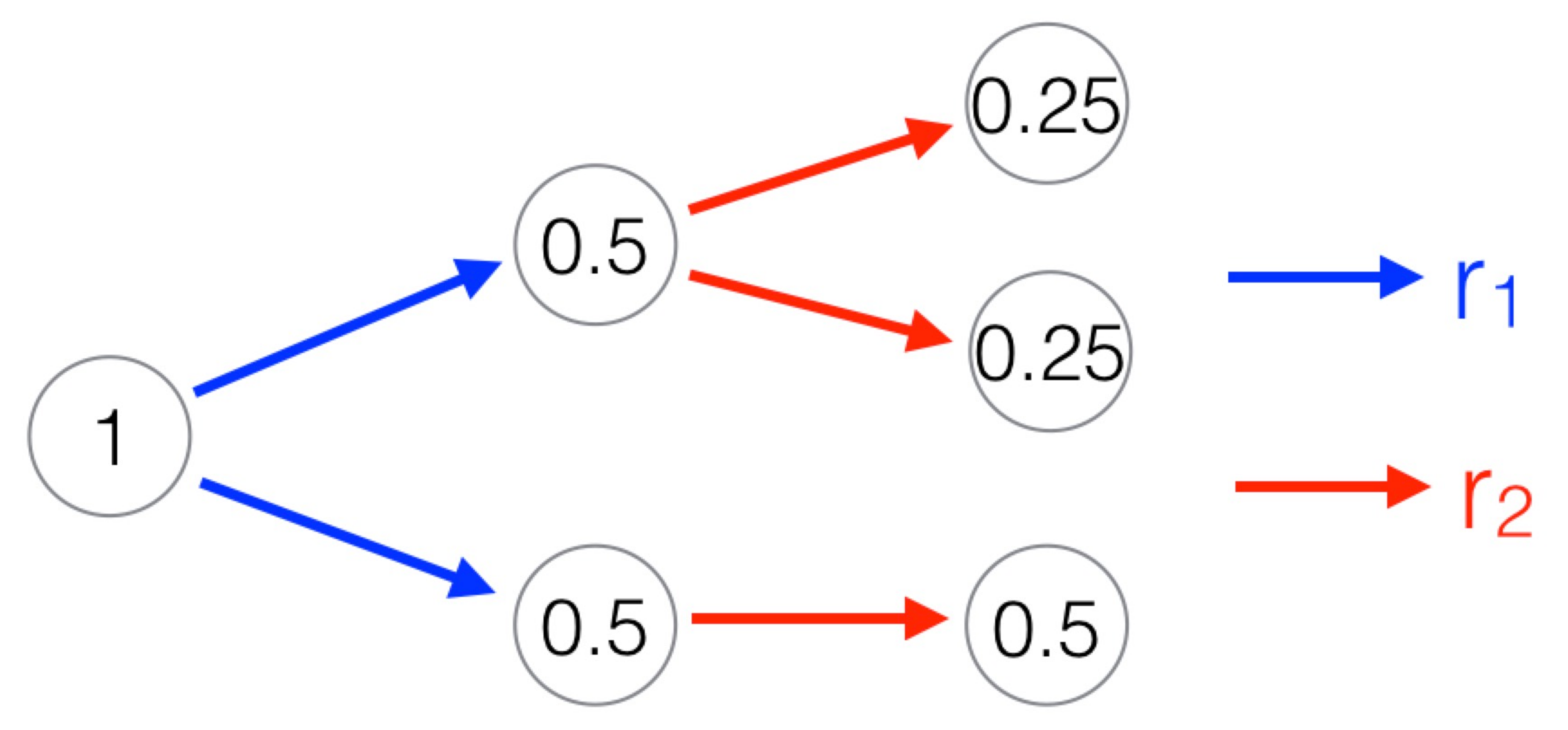}
	\caption{An illustration of the path-constraint resource algorithm (PCRA).}
	\label{fig:path}
\end{figure}

Although a knowledge base may contain a variety of relation paths between two entities, not every path is equally useful for inferring direct relations. For example, the relation path $John \xrightarrow{friend} Tim \xrightarrow{gender} male$ gives little contribution to inferring the gender of $John$.

 To evaluate the reliability of a path, PTransE uses a path-constraint resource algorithm (PCRA)~\cite{PTransE}, which is also applied in our approach. This algorithm first assigns a certain amount of resource (i.e., a value of 1) to the head entity $h$; then each node distributes resource evenly to its direct child nodes (Figure~\ref{fig:path}). The value of $p$ along an entity pair $h,t$ is denoted as $v(p|h,t)$.

However, $v(p|h,t)$ alone does not embody the  relatedness between a relation path $p$ and a direct relation $r$. To address this problem, a relatedness measure is defined as $P_r(r|p) = P_r(r,p)/P_r(p)$, where $P_r(p)$ is the sum of $v(p|h,t)$ for every training triplet ($h,r,t$) with $p$ as a relation path from $h$ to $t$. $P_r(r,p)$ is the sum of $v(p|h,t)$ for every training triplet ($h,r,t$) with $r$ being a direct relation and $p$ as a relation path.
The overall reliability of a path $p$ on a triplet ($h,r,t$) is given by
 
\begin{equation}
R(p|h,r,t) = P_r(r|p) \cdot v(p|h,t).
\end{equation}

PTransR's scoring function $f_\text{PTransR}(h,r,t)$ is composed of two scores:
\begin{equation}
	f_\text{PTransR}(h,r,t) = E(h,r,t) + E(\textbf{P}|h,r,t).
\end{equation}
$E(h,r,t)$ is the same as the scoring function of TransR (Equation~\ref{eqn:TransR})
which evaluates the plausibility of ($h,r,t$) without considering relation paths from $h$ to $t$.
Following PTransE, $E(\textbf{P}|h,r,t)$ is defined as 
\begin{equation}
E(\textbf{P}|h,r,t)=\dfrac{1}{Z} \sum_{p\in \textbf{P}} E(p|h,r,t),
\end{equation}
\begin{equation}
E(p|h,r,t) = R(p|h,r,t)\norm{ \textbf{p} - \textbf{r} }_2^2 = P_r(r|p) v(p|h,t)\norm{ \textbf{p} - \textbf{r} }_2^2,
\end{equation}
where $Z$ is a normalizing factor for $R(p|h,t)$ and $\textbf{P}$ is the set of all paths from $h$ to $t$. $E(\textbf{P}|h,r,t)$ evaluates the plausibility of ($h,r,t$) with the consideration of relation paths from $h$ to $t$. The overall loss function of PTransR is 
\begin{equation}
	\mathcal{L}_\text{PTransR}=\sum_{(h,r,t) \in S}[L(h,r,t) +  \dfrac{1}{Z}\sum_{p \in \textbf{P}}L(p|h,r,t)],
\end{equation}
\begin{equation}
	L(h,r,t) = \sum_{(h',r,t') \notin S} \max\big\{0,\gamma_1+E(h,r,t)-E(h',r,t')\big\},
\end{equation}
\begin{equation}
L(p|h,r,t) = \sum_{(h,r',t) \notin S} \max\big\{0,\gamma_2+E(p|h,r,t)-E(p|h,r',t)\big\},
\end{equation}
 PTransR learns entity and relation embeddings by minimizing $\mathcal{L}_\text{PTransR}$.

\subsection{Training Details}

\noindent We train PTransR by mainly following PTransE~\cite{PTransE}.

\noindent \textbf{Initial vectors and matrices.} Following TransR, initial vectors and matrices for PTransR are obtained from TransE. The configuration of TransE is: margin $\gamma= 1$, learning rate $\alpha= 0.01$, method = unif, and epoch = 1000.

\noindent\textbf{Negative samples.} We sample negative triplets by randomly replacing head entity $h$ or tail entity $t$ or relation $r$. For example, ($h,r,t$)-derived negative triplets are ($h',r,t$), ($h,r,t'$), and ($h,r',t$), where ($h,r,t$) $\in S$ and $(h',r,t)$, $(h,r',t)$, $(h,r,t')$$\notin S$.

\noindent\textbf{Vector representation constraints.} Following TransR, to regularize the representations, we impose the following constraints on the entity and relation embeddings.
\begin{equation}
	\parallel \textbf{h} \parallel = \parallel \textbf{t} \parallel = \parallel \textbf{r} \parallel = 1,\parallel \textbf{M}_{r}\textbf{h} \parallel \leq 1, \parallel \textbf{M}_{r}\textbf{t\textbf{}} \parallel \leq 1.
\end{equation}

\noindent\textbf{Path selection.} PTranE restricts the length of path to less than 3. Its results show that 3-hop paths do not make significant improvement, compared to 2-hop paths. For efficiency, we only consider 2-hop relation paths.

\noindent\textbf{Inverse relation.} As inverse relations sometimes contain useful information, for each training triplet ($h,r,t$), ($t,r^{-1},h$) is added to the training set.

\section{Evaluation}\label{sec:exp}

In this section, we present results of our experiment. We first briefly introduce the dataset and the task of entity prediction. Then we show the experimental results and analyze the performance.

\subsection{Dataset}\label{ss:dataset}

FB15k is a commonly used dataset in knowledge base completion. Table~\ref{tab:statistics} shows statistics of FB15k. FB15k dataset contains factual information in our world, e.g.,\texttt{ location/country/language\_spoken}. As FB15k has various kinds of relation, it is suitable for the evaluation of PTransR. Therefore, we choose FB15k as our experimental dataset.

\begin{table}[!t]
	\centering
	\begin{tabular}{|c|c|c|c|c|c|}
		\hline
		\makebox[2cm]{\textbf{Dataset}} &
		\makebox[2cm]{\textbf{Relation}} &
		\makebox[2cm]{\textbf{Entity}} & 
		\makebox[2cm]{\textbf{Train}} & 
		\makebox[2cm]{\textbf{Valid}} & 
		\makebox[2cm]{\textbf{Test}} \\
		\hline
		\textbf{FB15k} & 1,345 & 14,951 & 483,142 & 50,000 & 59,071\\
		\hline
	\end{tabular}
	
	%\centering
	\captionsetup{skip=5pt}
	
	\caption{Statistics of the FB15k dataset.}
	\label{tab:statistics}
\end{table}

\subsection{Experimental Settings}\label{ss:settings}

We evaluate PtransR on the task of entity prediction. Entity prediction aims at predicting the missing entity in an incomplete triplet, i.e., predicting $h$ given $r$ and $t$, or predicting $t$ given $h$ and $r$. Following the settings in TransE, for a triplet ($h,r,t$), we replace the head entity $h$ with every entity $e$ and compute the score of ($e,r,t$). Entity candidates are ranked according to their scores. We repeat the same process to predict the tail entity $t$. Then we use the two metrices in TransE to evaluate the performance: MeanRank (average rank of the expected entity) and Hits@10 (proportion of triplets whose head/tail entity is among top-10 in the ranking). However, there could be several entities that are plausible for the same incomplete triplet. The plausible entities which are ranked before $h$ or $t$ may cause underestimation of performance. One solution is to remove other plausible entities in the ranking, which is referred to as a \textit{filter}. In comparison, the results without removing other plausible entities are referred to as \textit{raw}. A good model should achieve low MeanRank and high Hits@10.

To utilize the inverse relation, instead of only using the score $f_\text{PTransR}(h,r,t)$, we use the sum of $f_\text{PTransR}(h,r,t)$ and $f_\text{PTransR}(t,r^{-1},h)$ to rank the candidates, i.e.,
\begin{equation}
	score(h,r,t) = f_\text{PTransR}(h,r,t) + f_\text{PTransR}(t,r^{-1},h).
\end{equation}

To accelerate the testing process, we use the reranking method in PTransE. We first rank all candidates according to their scores which are computed by the scoring function of TransR, which means that path information is not considered in the first ranking. Then we rerank the top-500 candidates according to the scores computed by the scoring function mentioned above, namely $score(h,r,t)$.

The configurations for experiments are given as follows: learning rate $\alpha$ for SGD among \{0.01,0.001,0.0001\}, dimension of enitity space $\mathbb{R}^k$ and relation space $\mathbb{R}^d$ between \{20,50\}, $\gamma _1$ and $\gamma_2$ among \{1,2,4\}, batch size $B$ among \{480,960,4800\}. The optimal configuration on valid set is $\alpha = 0.001$, $k=d=50$, $\gamma_1=\gamma_2=1$, and $B=4800$. The training process is limited to less than 500 epochs.

\subsection{Overall Performance}\label{ss:performance}

\begin{table}
	\centering
	\setlength{\abovecaptionskip}{4pt}
	\setlength{\belowcaptionskip}{-18pt}
	\begin{tabular}{|c|c c|c c|}
		\hline
		\multirow{2}{*}{Metric} & \multicolumn{2}{|c}{Mean Rank} &  \multicolumn{2}{|c|}{Hits@10(\%)}\\
		& Raw & Filter & Raw & Filter \\
		\hline
		Unstructured (Bordes et al. 2012) &	1,074&	979 & 4.5 & 6.3\\
		RESCAL (Nickel, Tresp, and Kriegel 2011) &	828&	683 & 28.4 &44.1\\
		SE (Bordes et al. 2011) &	273&	162 &28.8 &39.8\\
		SME (linear) (Bordes et al. 2012) &	274&	154 & 30.7 & 40.8\\
		SME (bilinear) (Bordes et al. 2012) &	284&	158 &31.3 & 41.3\\
		LFM (Jenatton et al. 2012) &	283&	164 & 26.0 &33.1\\
		TransE (Bordes et al. 2013) & 243 &125 & 34.9 & 47.1\\
		TransH (unif) (Wang et al. 2014) &211 &84 & 42.5 & 58.5\\
		TransH (bern) (Wang et al. 2014) &212 &87 & 45.7 &64.4\\
		TransR (unif) (Lin et al. 2015) & 226 & 78 & 43.8 &65.5\\
		TransR (bern) (Lin et al. 2015) & 198 & 77 & 48.2 &68.7 \\
		CTransR (unif) (Lin et al. 2015) & 233 & 82 & 44.0 & 66.3 \\
		CTransR (bern) (Lin et al. 2015) & 199 & 75 &48.4 & 70.2\\
		PTransE (2-hop) (Lin et al. 2015) & 200 & 54 & 51.8 & 83.4 \\
		PTransE (3-hop) (Lin et al. 2015) & 207 & 58 & 51.4 &\textbf{84.6}\\
		\hline
		PTransR (2-hop) & \textbf{171} & \textbf{47} &\textbf{53.0} &84.3 \\
		\hline
	\end{tabular}
	\caption{Evaluation results of entity prediction on FB15k.}
	\label{tab:result}
\end{table}

Table~\ref{tab:result} shows the experimental results. By comparing the results of PTransR with the results of previous models, we have the following main observations: (1) PTransR outperforms TransR on every metric to a large margin, which shows that path-augmented model can achieve better results than one-hop base model. (2) PTransR outperforms PTransE in MeanRank and is comparable to PTransE in Hits@10, which shows that path-agumented model with a better one-hop base model can achieve better performance.

\begin{table}
	\centering
	\setlength{\abovecaptionskip}{4pt}
	\setlength{\belowcaptionskip}{-18pt}
	\begin{tabular}{|c|c c c c|c c c c|}
		\hline
		Tasks &
		\multicolumn{4}{c}{Predicting Head (Hits@10)} &
		\multicolumn{4}{|c|}{Predicting Tail (Hits@10)} \\
		\hline
		{Relation Category} & 1-to-1 & 1-to-N& N-to-1 & N-to-N & 1-to-1 & 1-to-N& N-to-1 & N-to-N \\
		\hline
		Unstructured  & 34.5 & 2.5 & 6.1 & 6.6 & 34.3 & 4.2 & 1.9 & 6.6\\
		SE  &35.6 & 62.6 & 17.2 & 37.5 & 34.9 & 14.6 & 68.3 & 41.3\\
		SME (linear)  &35.1 & 53.7 & 19.0 & 40.3 & 32.7 & 14.9 & 61.6 & 43.3\\
		SME (bilinear)  &30.9 & 69.6 & 19.9 & 38.6 & 28.2 & 13.1 & 76.0 & 41.8\\
		TransE &43.7 & 65.7 & 18.2 & 47.2 & 43.7 & 19.7 & 66.7 & 50.0\\
		TransH (unif) & 66.7 & 81.7 & 30.2 & 57.4 & 63.7 & 30.1 &83.2 & 60.8\\
		TransH (bern) &66.8 & 87.6 &28.7 & 64.5 & 65.5 & 39.8 & 83.3 &67.2\\
		TransR (unif) &76.9 & 77.9 & 38.1 & 66.9 & 76.2 & 38.4 & 76.2 & 69.1\\
		TransR (bern) &78.8 & 89.2 & 34.1 & 69.2 & 79.2 & 37.4 & 90.4 & 72.1 \\
		CTransR (unif) &78.6 & 77.8 & 36.4 & 68.0 & 77.4 & 37.8 & 78.0 & 70.3 \\
		CTransR (bern) &81.5 & 89.0 &34.7 & 71.2 & 80.8 &38.6 & 90.1 & 73.8\\
		PTransE (2-hop) &91.0 & 92.8 & 60.9 & 83.8 & \textbf{91.2} & 74.0 & 88.9 & 86.4 \\
		PTransE (3-hop) &90.1 &92.0 &58.7 & \textbf{86.1} & 90.7 & 70.7 & 87.5 & \textbf{88.7}\\
		\hline
		PTransR (2-hop) &\textbf{91.4} & \textbf{93.4}& \textbf{65.5} & 84.2 & \textbf{91.2} & \textbf{74.5} & \textbf{91.8} & 86.8 \\
		\hline
	\end{tabular}
	\caption{Evaluation results of different relation catogories.}
	\label{tab:n2n}
\end{table}

Table~\ref{tab:n2n} presents the performance on the four relation catogories $1$-to-$1$, $1$-to-$n$, $n$-to-$1$, and $n$-to-$n$, with Hits@10(\textit{filter}) as the metric. From Table~\ref{tab:n2n}, we find that, compared to TransR, PTransR shows consistent imporvement on all four relation categories. Also, compared to PTransE, PTransR performs better on $1$-to-$1$, $1$-to-$n$ and $n$-to-$1$ relations, especially on $1$-to-$n$ and $n$-to-$1$.

\subsection{In-Depth Analysis and Discussion}

\begin{table}[!t]
	\centering
	\begin{tabular}{|c|c|c|c|c|c|}
		\hline
		Relation Frequency in Train Set & 1-3 & 4-15 & 16-50 & 51-300 & $>$300 \\
		\hline
		Relation Number & 291 & 305 & 243 & 271 & 235\\
		\hline
		MeanRank of TransR & 159 & 98 & 54 & 81 & 202 \\
		MeanRank of PTransR & 85 & 63 & 41 & 63 & 182 \\
		improvement (\%) & 46.5 & 35.7 & 24.1 & 22.2 & 9.9 \\		
		\hline
	\end{tabular}\vspace{.3cm}
	\caption{Evaluation results concerning relations of different frequency in train set.}
	\label{tab:sparsity}
\end{table}

As pointed out in Section~\ref{sec:intro}, despite the massive train set of FB15k, some relations cannot be properly captured due to the problem of data sparsity. We separate relations into five groups according to their frequency in the train set, as shown in Table~\ref{tab:sparsity}. MeanRank(\textit{raw}) of TransR and PTransR is compared in Table~\ref{tab:sparsity} and the improvement from TransR to PTransR is presented. First of all, we see PTransR outperforms TransR in all five groups of relations. Second, as relation frequency decreases, the improvement goes up, which means that path information is useful for dealing with the problem of data sparsity.

\section{Related Work}\label{sec:relatedwork}

Relation extraction is an important research topic in NLP. It can be roughly divided into two categories based on the source of information. 

Text-based approaches extraction entities and/or relations from plain text. For example, Hearst~\cite{hearst} uses ``is a$|$an'' pattern to extract hyponymy relations. Banko et al.~\cite{openIE} proposes to extract open-domain relations from the Web. Fully supervised relation extraction, which classify two marked entities into several predefined relations, have become a hot research arena in the past several years \cite{santos,Xu,SDP-LSTM}.

Knowledge base completion/population, on the other hand, does not use additional text. 
Socher et al.~\cite{socher} propose a tensor model to predict missing relations in an existing knowledge base, showing neural networks' ability of entity-relation inference. 
Then, translating embeddings approaches are proposed for knowledge base completion \cite{TransE,TransH,TransR,PTransE}.
Recently, Wang et al.~\cite{juanzi} use additional information to improve knowledge base completion by using textual context.

In this paper, we focus on pure knowledge base completion, i.e., we do not use additional resources. We combine the state-of-the-art one-hop TransR model~\cite{TransR} and path augmentation method~\cite{PTransE}, resulting in the new PTransR variant.

\section{Conclusion}\label{sec:conclusion}

In this paper, we augment one-hop TransR model with path modeling method, resulting in PTransR model. We evaluate PTransR on the task of entity prediction and compare the performance of PTransR with that of previous models. Experimental results show that path information is useful in sovling the problem of data sparsity, and that PTransR outperforms previous models, which makes PTransR the state-of-the-art model in the field that populates knowledge base without using additional text.

\bibliographystyle{splncs}
\bibliography{kk}

\begin{thebibliography}{10}

\bibitem{wordnet}
Miller, G.A.:
\newblock {WordNet: A lexical database for English}.
\newblock Communications of the ACM \textbf{38}(11) (1995)  39--41

\bibitem{freebase}
Bollacker, K., Evans, C., Paritosh, P., Sturge, T., Taylor, J.:
\newblock {Freebase: A} collaboratively created graph database for structuring
  human knowledge.
\newblock In: Proceedings of the 2008 ACM SIGMOD International Conference on
  Management of Data. (2008)  1247--1250

\bibitem{websearch}
Jiang, X., Tan, A.H.:
\newblock Learning and inferencing in user ontology for personalized semantic
  web search.
\newblock Information Sciences \textbf{179}(16) (2009)  2794--2808

\bibitem{QA}
Yao, X., Van~Durme, B.:
\newblock Information extraction over structured data: Question answering with
  freebase.
\newblock In: Proceedings of the 52nd Annual Meeting of the Association for
  Computational Linguistics. (2014)  956--966

\bibitem{gene}
Ashburner, M., Ball, C.A., Blake, J.A., Botstein, D., Butler, H., Cherry, J.M.,
  Davis, A.P., Dolinski, K., Dwight, S.S., Eppig, J.T.,  et~al.:
\newblock {Gene Ontology: T}ool for the unification of biology.
\newblock Nature Genetics \textbf{25}(1) (2000)  25--29

\bibitem{openIE}
Banko, M., Cafarella, M.J., Soderland, S., Broadhead, M., Etzioni, O.:
\newblock Open information extraction from the web.
\newblock In: Proceedings of the 20th International Joint Conference on
  Artificial Intelligence. (2007)  2670--2676

\bibitem{TransH}
Wang, Z., Zhang, J., Feng, J., Chen, Z.:
\newblock Knowledge graph embedding by translating on hyperplanes.
\newblock In: Proceedings of the 28th AAAI Conference on Artificial
  Intelligence. (2014)  1112--1119

\bibitem{SDP-LSTM}
Xu, Y., Mou, L., Li, G., Chen, Y., Peng, H., Jin, Z.:
\newblock Classifying relations via long short term memory networks along
  shortest dependency paths.
\newblock In: Proceedings of the Conference on Empirical Methods in Natural
  Language Processing. (2015)  1785--1794

\bibitem{webIE}
Chang, C.H., Kayed, M., Girgis, M.R., Shaalan, K.F.:
\newblock A survey of {W}eb information extraction systems.
\newblock IEEE Transactions on Knowledge and Data Engineering \textbf{18}(10)
  (2006)  1411--1428

\bibitem{wiki}
Zesch, T., M{\"u}ller, C., Gurevych, I.:
\newblock Extracting lexical semantic knowledge from {Wikipedia and
  Wiktionary}.
\newblock In: Proceedings of the 6th International Conference on Language
  Resources and Evaluation. (2008)  1646--1652

\bibitem{TransE}
Bordes, A., Usunier, N., Garcia-Duran, A., Weston, J., Yakhnenko, O.:
\newblock Translating embeddings for modeling multi-relational data.
\newblock In: Advances in Neural Information Processing Systems. (2013)
  2787--2795

\bibitem{TransR}
Lin, Y., Liu, Z., Sun, M., Liu, Y., Zhu, X.:
\newblock Learning entity and relation embeddings for knowledge graph
  completion.
\newblock In: Proceedings of the 29th AAAI Conference on Artificial
  Intelligence. (2015)  2181--2187

\bibitem{liang}
Guu, K., Miller, J., Liang, P.:
\newblock Traversing knowledge graphs in vector space.
\newblock In: Proceedings of the 2015 Conference on Empirical Methods in
  Natural Language Processing, Lisbon, Portugal, Association for Computational
  Linguistics (September 2015)  318--327

\bibitem{PTransE}
Lin, Y., Liu, Z., Luan, H., Sun, M., Rao, S., Liu, S.:
\newblock Modeling relation paths for representation learning of knowledge
  bases.
\newblock In: Proceedings of the Conference on Empirical Methods in Natural
  Language Processing. (2015)  705--714

\bibitem{analogy}
Mikolov, T., Yih, W.T., Zweig, G.:
\newblock Linguistic regularities in continuous space word representations.
\newblock In: Proceedings of the 2013 Conference of the North American Chapter
  of the Association for Computational Linguistics: Human Language
  Technologies, Association for Computational Linguistics (2013)  746--751

\bibitem{SNLI}
Mou, L., Men, R., Li, G., Xu, Y., Zhang, L., Yan, R., Jin, Z.:
\newblock Natural language inference by tree-based convolution and heuristic
  matching.
\newblock In: The 54th Annual Meeting of the Association for Computational
  Linguistics. (2016)  130

\bibitem{hearst}
Hearst, M.A.:
\newblock Automatic acquisition of hyponyms from large text corpora.
\newblock In: Proceedings of the 14th conference on Computational
  linguistics-Volume 2, Association for Computational Linguistics (1992)
  539--545

\bibitem{santos}
dos Santos, C.N., Xiang, B., Zhou, B.:
\newblock Classifying relations by ranking with convolutional neural networks.
\newblock arXiv preprint arXiv:1504.06580 (2015)

\bibitem{Xu}
Xu, Y., Jia, R., Mou, L., Li, G., Chen, Y., Lu, Y., Jin, Z.:
\newblock Improved relation classification by deep recurrent neural networks
  with data augmentation.
\newblock In: COLING. (2016)

\bibitem{socher}
Socher, R., Chen, D., Manning, C.D., Ng, A.:
\newblock Reasoning with neural tensor networks for knowledge base completion.
\newblock In: Advances in Neural Information Processing Systems. (2013)
  926--934

\bibitem{juanzi}
Wang, Z., Li, J.:
\newblock Text-enhanced representation learning for knowledge graph.
\newblock In: International Joint Conference on Artificial Intelligence. (2016)
   1293--1299

\end{thebibliography}

\end{document}